\documentclass[sigconf, authordraft]{acmart}
\AtBeginDocument{%
  }

\usepackage{xspace}
\usepackage{bbm}
\usepackage{caption}
\usepackage{wrapfig}
\usepackage{adjustbox}
\usepackage{subcaption}

\newcommand{\eg}[1]{\emph{e.g.}}

\newcommand{\method}{Stereo-GS\xspace}

\newcommand{\Tref}[1]{Table~\ref{#1}}

\newcommand{\Fref}[1]{Figure~\ref{#1}}

\setcopyright{acmlicensed}
\copyrightyear{2025}
\acmYear{2025}
\setcopyright{cc}
\setcctype{by}
\acmConference[MM '25]{Proceedings of the 33rd ACM International Conference on Multimedia}{October 27--31, 2025}{Dublin, Ireland}
\acmBooktitle{Proceedings of the 33rd ACM International Conference on Multimedia (MM '25), October 27--31, 2025, Dublin, Ireland}
\acmDOI{10.1145/3746027.3755151}
\acmISBN{979-8-4007-2035-2/2025/10}

\begin{document}


\title{Stereo-GS: Multi-View Stereo Vision Model for Generalizable 3D Gaussian Splatting Reconstruction}

\author{Xiufeng Huang}
\email{xiufenghuang@life.hkbu.edu.hk}
\affiliation{%
  \institution{Department of Computer Science, Hong Kong Baptist University}
  \city{Hong Kong}
  \country{China}
}

\author{Ka Chun Cheung}
\email{chcheung@nvidia.com}
\affiliation{%
  \institution{NVIDIA AI Technology Center, NVIDIA}
  \city{Hong Kong}
  \country{China}
}

\author{Runmin Cong}
\email{rmcong@sdu.edu.cn}
\affiliation{%
  \institution{School of Control Science and
  Engineering,
  Shandong University}
  \city{Jinan}
  \state{Shandong}
  \country{China}
}

\author{Simon See}
\email{ssee@nvidia.com}
\affiliation{%
 \institution{NVIDIA AI Technology Center, NVIDIA}
 \city{Singapore}
 \country{Singapore}
}

\author{Renjie Wan}
\authornote{Corresponding author. This work was carried out at the Renjie Group, Hong Kong Baptist University.}
\email{renjiewan@hkbu.edu.hk}
\affiliation{%
  \institution{Department of Computer Science, Hong Kong Baptist University}
  \city{Hong Kong}
  \country{China}
}

\renewcommand{\shortauthors}{Xiufeng Huang et al.}

\begin{abstract}

Generalizable 3D Gaussian Splatting reconstruction showcases advanced Image-to-3D content creation but requires substantial computational resources and large datasets, posing challenges to training models from scratch. 
Current methods usually entangle the prediction of 3D Gaussian geometry and appearance, which rely heavily on data-driven priors and result in slow regression speeds.
To address this, we propose \method, a disentangled framework for efficient 3D Gaussian prediction. 
Our method extracts features from local image pairs using a stereo vision backbone and fuses them via global attention blocks. 
Dedicated point and Gaussian prediction heads generate multi-view point-maps for geometry and Gaussian features for appearance, combined as GS-maps to represent the 3DGS object. 
Unlike existing methods that require camera parameters, our approach reconstructs 3D Gaussians from images without camera dependency.
A subsequent refinement network enhances the predicted GS-maps, optionally conditioned on camera poses.
By reducing resource demands while maintaining high-quality outputs, \method provides an efficient, scalable solution for real-world 3D content generation.
Project page: https://kevinhuangxf.github.io/stereo-gs.

\end{abstract}



\begin{teaserfigure}
  \includegraphics[width=\textwidth]{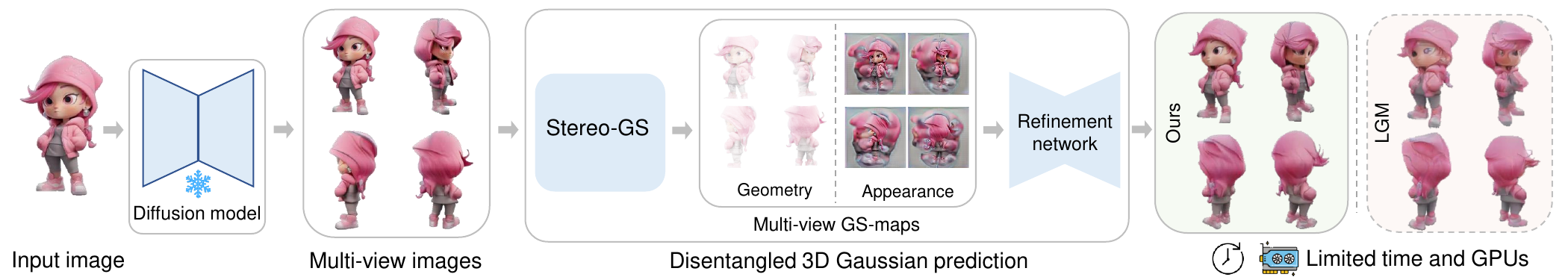}
  \caption{We propose \method for efficient training and inference of generalizable 3D Gaussian Splatting (3DGS) \cite{kerbl3Dgaussians} reconstruction for Image-to-3D generation. Given a single input image, we use a diffusion model to generate multi-view images. Our \method then employs a novel multi-view stereo vision model to extract features from these images, predicting 3DGS geometry (point-maps) and appearance in a disentangled manner to construct multi-view GS-maps as a representation of 3DGS objects. Unlike existing methods \cite{hong2023lrm, tang2024lgm} such as LGM \cite{tang2024lgm}, which produce degraded results with artifacts under limited resources, our approach generates high-quality 3DGS content even with constrained training resources.}
  \Description{\method teasear figure.}
  \label{fig:teaser}
\end{teaserfigure}


\maketitle

\section{Introduction}
\label{sec:introduction}

Images-to-3D has become a crucial step in 3D content generation, with numerous studies focused on enhancing the quality of generated content \cite{hong2023lrm, szymanowicz2024splatter, zou2024triplane, tang2024lgm}. However, the high demand for training resources poses a challenge for creators seeking to train models from scratch. How can we reduce these training resource demands while still maintaining high-quality generation?

The major training resources for Image-to-3D approaches \cite{hong2023lrm, tang2024lgm, xu2024grm} consist of computing power and datasets. To maintain high generation quality, current approaches typically rely on substantial training resources, including large-scale datasets and high-end GPUs, and resulting in significant amounts of training time. When training resources are insufficient, current approaches often experience a severe drop in performance. For example, with fewer training resources, LRM \cite{hong2023lrm} can only render images at a resolution of $128 \times 128$ size. LGM \cite{tang2024lgm} also suffers from blurry rendering and coarse geometry results due to the time-consuming regression.


The primary reason for this dilemma stems from the current methods’ entanglement of 3DGS geometry and appearance predictions \cite{kerbl3Dgaussians, tang2024lgm, xu2024grm}, which forces models to rely heavily on data-driven priors to achieve generalizability. These priors, learned from large-scale datasets and extensive computing resources, enable 3D reconstruction from sparse or single images. 
However, as shown in \Fref{fig:teaser}, limited training resources lead to significant performance degradation in 3D reconstruction models. 
This entanglement forces models to depend heavily on data-driven priors for generalization, which become unreliable under resource constraints. Consequently, decoupling geometry-appearance predictions to reduce reliance on data-driven priors is critical to improving training efficiency and robustness.


To reduce the data-driven reliance, a more reasonable way is to consider a disentangled framework that separates the 3D Gaussian prediction into two distinct tasks: the 3D reconstruction of point clouds for geometry and the prediction of Gaussian features for appearance. 
This disentangled framework allows us to integrate pre-trained deep neural networks as geometry priors~\cite{williams2019deep} to accelerate 3D reconstruction training.
Subsequently, the appearance parameter optimization benefits significantly from the accurate geometric conditions already established, enabling more efficient convergence.
For example, Triplane-Gaussian \cite{zou2024triplane} uses Point-E \cite{nichol2022point} to generate 4096-point clouds from a single image, generating as the geometry prior. These point clouds are then decoded into Gaussian features via a tri-plane decoder to model appearance.
However, Point-E \cite{nichol2022point} can only generate sparse point clouds ($4096$ points) from a single image and result in undermined 3D reconstruction results. 
To avoid sparse point clouds prediction, we explore stereo vision models to predict dense point clouds for improved reconstruction quality.

Recent stereo vision models \cite{dust3r_cvpr24, mast3r_eccv24} showcase outstanding geometry reconstruction performance from multiple perspective images. 
Particularly, DUSt3R \cite{dust3r_cvpr24} can generate dense 3D point clouds with precise geometry by building pixel-wise 2D-to-3D mappings, and MASt3R \cite{mast3r_eccv24} can further predict matching features for finding correspondences across multiple images.
The 3D point clouds and matching features can serve as the geometry and matching priors to accelerate the 3D reconstruction model training.
However, the 3D point clouds from multi-view images can not be directly fused together, since these stereo vision models \cite{dust3r_cvpr24, mast3r_eccv24} are only generalizable in the local pair-wise views. It relies on per-scene optimization to align the point clouds globally, which is time-consuming and impractical for 3D reconstruction in a feed-forward manner.

\begin{figure*}[tp]
  \centering
  \begin{subfigure}{\linewidth}
    \includegraphics[width=1.0\linewidth]{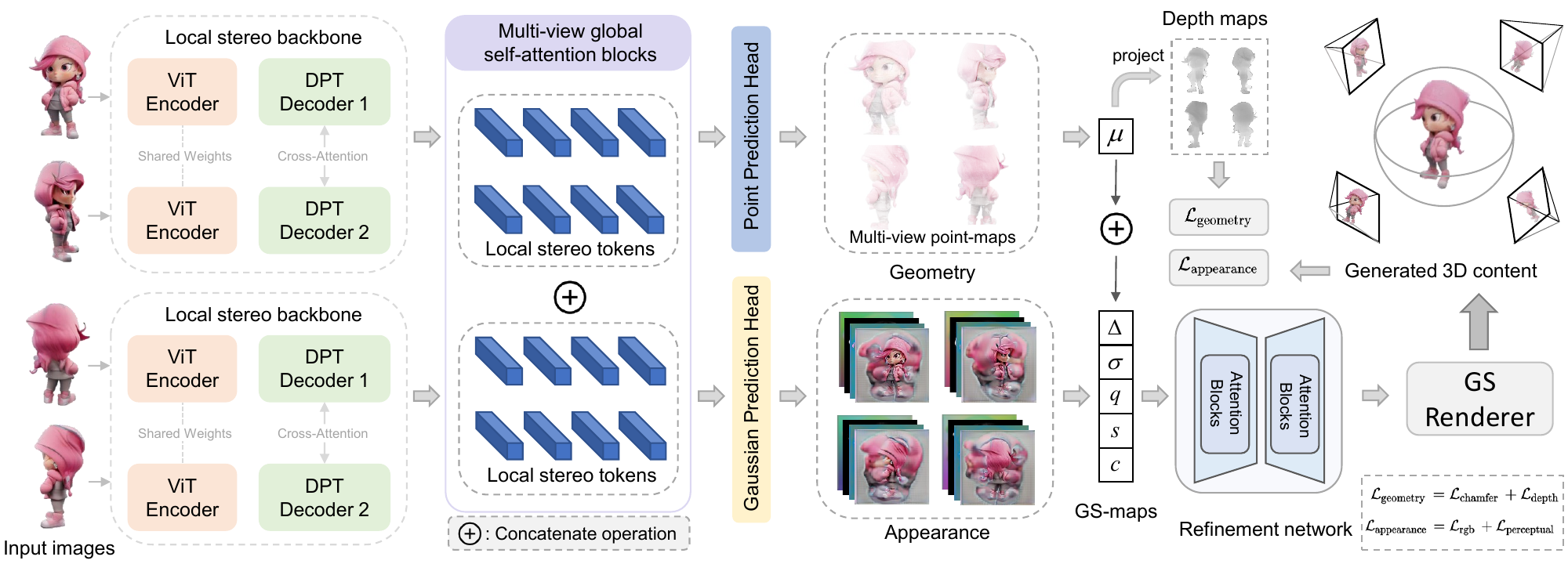}
  \end{subfigure}
  
  \caption{
  Our proposed \method generates multi-view GS-maps in a disentangled manner for predicting 3DGS geometry and appearance, enabling high-quality 3D Gaussian reconstruction. It first uses a stereo vision model to extract local feature tokens from image pairs, which are fused via multi-view global attention blocks. A point prediction head estimates geometry through multi-view point-maps, while a Gaussian prediction head generates Gaussian features for appearance. These are combined into GS-maps representing the 3DGS object, refined by a cross-view attention-based network, and rendered as per-pixel 3D Gaussians for novel views during training.
  }
  \label{fig:algo_overview}
\end{figure*}

We propose \method to extend these stereo vision models \cite{dust3r_cvpr24, mast3r_eccv24} from local stereo setups into a multi-view stereo framework for generalizable 3D Gaussian splatting (3DGS) reconstruction \cite{kerbl3Dgaussians, tang2024lgm}.
As shown in \Fref{fig:algo_overview}, our method employs a stereo vision model as a backbone to extract features from local image pairs. These features are then fused using multi-view global attention blocks, enabling the disentangled generation of 3DGS geometry and appearance.
Based on the fused multi-view stereo tokens, we design a point prediction head to predict point-maps for geometry and a Gaussian prediction head to predict Gaussian features for appearance.
The point-maps and Gaussian features are then combined as the multi-view GS-maps for representing the 3DGS objects.
Such a disentangled design for 3DGS geometry and appearance ensures high efficiency in training time.

Current 3D Gaussian reconstruction models \cite{tang2024lgm, xu2024grm, gslrm2024} often rely on camera parameters, which can limit their applicability in real-world scenarios. 
In contrast, our method leverages the unconstrained stereo vision models \cite{dust3r_cvpr24, mast3r_eccv24} to achieve generalizable 3D Gaussian reconstruction, ensuring high efficiency in inference time and enhancing practicality for real-world applications.
In summary, our contributions are listed below:
\begin{enumerate}
\item We propose \method, a novel method that extends stereo vision models \cite{dust3r_cvpr24, mast3r_eccv24} from local pairwise setups to a multi-view stereo framework.
By leveraging multi-view global attention blocks with point and Gaussian prediction heads, our method can generate precise 3D point-maps and Gaussian features in a feed-forward manner.
\item We introduce a disentangled training framework that separates to optimize 3DGS geometry and appearance. 
This disentanglement reduces the reliance on data-driven priors, improving training efficiency and robustness, especially under limited resources.
\item Our method reconstructs 3D Gaussians without camera dependency by leveraging unconstrained stereo feature backbones.
This design ensures robustness and practicality for real-world applications where camera data may be unavailable or unreliable.
\end{enumerate}



\section{Related work}
\label{sec:related_word}

\paragraph{Radiance field for novel view synthesis.} 

Novel View Synthesis (NVS) aims to generate new, unseen perspectives of an object or scene from a provided set of images, by creating 3D representations \cite{mildenhall2019local}.
Neural Radiance Fields \cite{mildenhall2021nerf, luo2025nerf, huang2024geometrysticker} (NeRF) have achieved photo-realistic representations of 3D scenes, encoded by multilayer perception (MLP) networks~\cite{mildenhall2021nerf, mueller2022instant, barron2022mip}. 
Recently, 3D Gaussian Splatting~\cite{kerbl3Dgaussians, song2024geometry, huang2024gaussianmarker, li2024variational} (3DGS) has demonstrated remarkable results by rendering each point in the 3D space via an efficient rasterization process.
NVS methods \cite{mildenhall2021nerf, kerbl3Dgaussians, song2024protecting, luo2023copyrnerf} usually require scene-specific optimization, resulting in increased computational demands and longer processing times.
Compared with per-scene optimization methods, feed-forward-based methods~\cite{szymanowicz2024splatter, wang2021ibrnet, chen2024mvsplat} can achieve generalizable 3D reconstruction on the entire scene in a single pass, eliminating the need for additional optimization.

\paragraph{Generalizable 3D reconstruction.} 

With advancements in the radiance field, there is a growing interest in generalizable 3D reconstruction from multi-view images \cite{ni2024colnerf, yu2021pixelnerf, shi2024zerorf}, which may contain limited overlaps.
Specifically, unconstrained 3D reconstruction methods~\citep{ye2024no, xu2024sparp, wang2023pf, lin2024relpose++} are attracting significant interest due to their generalization and pose-free capabilities.
Recent unconstrained stereo vision frameworks DUSt3R \cite{dust3r_cvpr24} and MASt3R \cite{mast3r_eccv24} showcase outstanding 3D reconstruction performance from multi-view images.
DUSt3R \cite{dust3r_cvpr24} can generate 3D point clouds from pair-wise images by building the 2D-to-3D correspondences. 
MASt3R \cite{mast3r_eccv24} can further predict matching features between the pair-wise views. 
However, both DUSt3R \cite{dust3r_cvpr24} and MASt3R \cite{mast3r_eccv24} are only generalizable in local pair-wise views and require post-hoc alignment to merge multi-view point clouds.
This motivates us to lift the stereo models from a local pair-wise setting into a multi-view stereo framework.

\paragraph{Multi-view diffusion models.}

2D diffusion models~\cite{rombach2022high, saharia2022photorealistic} are initially designed for generating single-view images without the ability for novel view synthesis. 
The recent progress on multi-view diffusion model~\cite{wang2023imagedream, shi2023mvdream, xu2023dmv3d, huang2024mv} empowers the 2D diffusion model to generate multi-view perspectives for 3D objects, by leveraging the camera poses as the conditional input. 
However, the Multi-view diffusion model can usually generate only a few views with less than 6 images, and inconsistencies may still occur across the generated views. 
Recently, video diffusion models \cite{chen2024v3d, voleti2025sv3d} have demonstrated the capability for object-centric multi-view generation, producing more output frames and achieving even more consistent results. Leveraging the power of multi-view diffusion models or video diffusion models \cite{chen2024v3d, voleti2025sv3d}, we can accomplish Image-to-3D and Text-to-3D tasks. These models generate multi-view images, which serve as inputs for our proposed method to reconstruct 3D objects.


\section{Preliminary of 3DGS}
\label{sec:preliminary}

Starting from a sparse set of Structure-from-Motion (SfM) \cite{snavely2006photo} points, the goal of 3DGS \cite{kerbl3Dgaussians} is to optimize a scene representation that enables high-quality novel view synthesis. 
The scene is modeled as a collection of 3D Gaussians, which represent the radiance emitted in the 3D space around each point. 
Each 3D Gaussian is parameterized by its mean $\mu \in \mathbb{R}^3$ as the position, opacity $\alpha \in \mathbb{R}^3$ for transparency, Gaussian covariance matrix $\Sigma$ as she shape, and spherical harmonics $c \in \mathbb{R}^{3 \times d}$ as the view-dependent color.
By utilizing a scaling matrix $S$ and rotation matrix $R$, we can determine the corresponding $\Sigma=RSS^{T}R^{T}$ and ensure $\Sigma$ is positive semi-definite.
In our experiments, we focus on view-independent RGB color $S \in \mathbb{R}^{3}$ in the spherical harmonics for each Gaussian.
The 3D Gaussians need to be further projected to 2D Gaussians for rendering by volume splatting \cite{zwicker2001ewa} method.
During rendering, 3DGS follows a typical neural point-based approach \cite{kopanas2022neural} to compute the color $C$ of a pixel by blending $\mathcal{N}$ depth ordered points:
\begin{equation}
C=\sum_{i \in \mathcal{N}} c_i \alpha_i \prod_{j=1}^{i-1}\left(1-\alpha_j\right),
\label{eq:3dgs_rendering}
\end{equation}
where $c_i$ is the color estimated by the spherical harmonics (SH) coefficients of each Gaussian, and $\alpha_i$ is given by evaluating a 2D Gaussian with covariance $\Sigma^{'}$ \cite{yifan2019differentiable} multiplied with a per-point opacity.

\section{Proposed method}
\label{sec:method}



We introduce a novel Images-to-3D framework, as illustrated in \Fref{fig:algo_overview}. 
First, we leverage a pre-trained diffusion model \cite{chen2024v3d} to generate multi-view images from a single input image. 
Then, we employ our proposed \method model to estimate multi-view point-maps for geometry and Gaussian features for appearance. 
Finally, these outputs are combined as GS-maps and enhanced by a refinement network, ensuring the generation of high-quality 3D Gaussians that accurately represent the 3D objects.

\subsection{Stereo-GS}

Current unconstrained stereo vision models, such as DUSt3R \cite{dust3r_cvpr24} and MASt3R \cite{mast3r_eccv24}, are typically designed for local image pairs, while rely on a time-consuming fine-tuning based global alignment for multi-view stereo vision.
To address this limitation, we propose a novel method to lift these stereo vision models from a two-view setting into a multi-view stereo setting, enabling efficient and accurate 3D reconstruction in a feed-forward manner.

\paragraph{Local stereo feature backbone}

Given overlapped local image pairs, stereo vision models like DUSt3R \cite{dust3r_cvpr24} and MASt3R \cite{mast3r_eccv24} can predict 3D point maps in local camera coordinates.
It first encodes both images with a Siamese ViT \cite{dosovitskiy2020vit} encoder $\operatorname{Enc}$, which yields two token representations $T^1$ and $T^2$. 
Then, a pair of twined ViT decoders $\operatorname{Dec^1}, \operatorname{Dec^2}$ sequentially performs cross-attention to the tokens with spatial information:
\begin{equation}
\begin{aligned}
    T^1 &= \operatorname{Enc}\left(I^1\right), \quad & T^2 &= \operatorname{Enc}\left(I^2\right), \\
    T_i^1 &= \text{Dec}_i^1\left(T_{i-1}^1, T_{i-1}^2\right), \quad & T_i^2 &= \operatorname{Dec}_i^2\left(T_{i-1}^2, T_{i-1}^1\right),
\end{aligned}
\end{equation}
where $\operatorname{Dec}_i^v\left(T^1, T^2\right)$ denotes the $i$-th block in local input view $v \in\{1,2\}$ with the input tokens $T^1, T^2$, and $i=1, \ldots, K$ for a decoder with $K$ blocks and initialized with encoder tokens $T_0^1:=T^1$ and $T_0^2:=T^2$.
Finally, two Dense Prediction Transformer (DPT) \cite{Ranftl2021dpt} heads predict the final pixel-wise point-maps $P$.
However, these stereo point-maps are in a local camera coordinate system, with only the possibility of post-processing alignment to merge the multiple local stereo point-maps.

Thus, we propose to utilize the stereo vision model as the local backbone to extract stereo feature tokens and reconstruct 3D objects within a unified world coordinate in a feed-forward manner.
Unlike current methods that rely on a separated DPT head \cite{Ranftl2021dpt} for predicting individual point-maps for each view, we utilize every stereo pair tokens $\{T^1_i, T^2_i\}$ as local features, and then directly concatenating them as multi-view stereo tokens to a multi-view attention head for the global fusion of 3D point-maps in a feed-forward manner:
\begin{equation}
\tilde{T}_i = \text{concat}\left(T_i^1, T_i^2, \dots, T_i^n\right),
\end{equation}
where $\tilde{T}_i$ represents the concatenated multi-view stereo tokens from $n$ input multi-view images.


\paragraph{Multi-view global attention blocks}


Current methods \cite{dust3r_cvpr24, mast3r_eccv24} apply separated DPT heads on the learned tokens for each input view image.
We propose multi-view self-attention blocks to learn global representation based on the multi-view stereo tokens $\tilde{T}$.
We select tokens at the layer of $i=\{3,6,9,12\}$ to resemble features at four different stages. 
These tokens are then fed into the feature fusion blocks \cite{Ranftl2021dpt} in the DPT head.
The selected tokens $\tilde{T}_i$ are subsequently processed by multi-view self-attention blocks:
\begin{equation}
    SA\left(\tilde{T}_i\right)=\operatorname{softmax}\left(\frac{Q \cdot K^T}{\sqrt{d_k}}\right) \cdot V.
\end{equation}
The output of the self-attention blocks is added to the input multi-view stereo tokens with a residual connection.
Since pair-wise cross-attention already attends to the tokens $T^1_i, T^2_i$ from the local stereo model, we only apply global self-attention attends to the muti-view stereo tokens $\tilde{T}_i$.
This design strikes a balance between efficiently integrating information across multi-view images globally and applying cross-attention on stereo image pairs locally. 

\paragraph{Point and Gaussian prediction head}


We incorporate a point prediction head to generate multi-view point-maps $P$, which represent the geometry of 3DGS objects. 
This head is designed similarly to the DPT head \cite{Ranftl2021dpt}, utilizing four feature fusion blocks that progressively upsample the multi-view tokens $\tilde{T}_i$ by a factor of two at each stage. 
The output layer consists of three convolutional layers: the first upsamples the feature predictions to match the input image resolution, while the last reduces the feature dimensions to three channels for point-map prediction. 
The final point-map representation is initially at half the input image resolution and is then interpolated to match the input image size.

Similarly, the Gaussian prediction head follows the same design as the point prediction head. 
It takes multi-view stereo tokens $\tilde{T}_i$ as input and generates an output feature map with $11$ channels, representing the Gaussian features for 3DGS appearance.


\subsection{Disentangled 3D Gaussians prediction}

Existing methods \cite{tang2024lgm, xu2024grm, gslrm2024} for generalizable 3D Gaussian prediction typically take multi-view images as input and entangle the prediction of 3D Gaussian geometry with other appearance parameters. 
Nonetheless, the absence of geometry prior knowledge leads to predicting the 3D Gaussian positions and learning multi-view correspondence reasoning in an implicit manner.
Such implicit predictions can cause the regression to solely rely on data-driven priors for learning multi-view correspondence reasoning and geometry estimation, which is time-consuming and can result in unreliable 3D Gaussian predictions.

We propose a disentangled two-stage training approach to predict 3D Gaussians from multi-view images, where the geometry and appearance of the 3D Gaussians are learned separately.
Unlike previous methods \cite{szymanowicz2024splatter, zou2024triplane, tang2024lgm}, which are limited to fixed input configurations, our method is flexible and can handle arbitrary multi-view images, significantly broadening its applicability.


\paragraph{Geometry reconstruction.}
In the first stage, we predict the 3DGS geometry using our proposed \method model by predicting the multi-view point-maps. 
These point-maps serve as an intermediate representation, effectively acting as a point cloud representation for the 3D Gaussian position parameters. 
By providing a strong geometric prior, our method accelerates the initial stage of 3D Gaussian prediction, enabling faster and more accurate reconstructions.

A key advantage of our \method is its pose-free nature, which ensures generalizability and robustness for direct 3D reconstruction tasks. In contrast, existing methods \cite{tang2024lgm} typically assume the availability of precise camera parameters during testing. This reliance on camera parameters not only limits their applicability to real-world scenarios but also makes them highly sensitive to inaccuracies in the provided parameters. For example, imperfect or noisy camera poses, especially in synthetic multi-view image settings, can lead to suboptimal reconstructions and degraded results.


To address these limitations, we train our \method model on large-scale datasets without relying on camera parameters, achieving a pose-free design in a world coordinate system. This approach ensures that our method is practical and reliable for real-world applications where camera data may be unavailable or unreliable.

\vspace{-0.5mm}

\paragraph{Appearance predictions.}


In the second stage, we train our model with the Gaussian prediction head for 3DGS appearance prediction. 
Each pixel in the output feature map represents a 3D Gaussian appearance attribute, similar to SplatterImage \cite{szymanowicz2024splatter}.
The output feature map consists of $11$ channels, encoding the appearance information of 3D Gaussians, including scaling $s$, rotation $q$, opacity $\sigma$ and RGB color $c$. We concatenate these appearance attributes with the previously predicted point maps $P$, which represent the 3D Gaussian positions $\mu$, to construct GS-maps that align with the standard 3D Gaussian parameters.

For the output GS-maps, we clamp the point-maps position $P$ into $[-1.0,1.0]^3$. 
Color $c$ and opacity $\alpha$ are activated with the sigmoid function. Rotation $r$ is $\ell_2$ normalized. 
Scaling $s$ is activated with the softplus function and multiplied with a small coefficient $0.1$.
We transform each GS-map into a set of Gaussians and directly merge them as the final 3D Gaussians, which are used to render images as novel views during training.

\paragraph{Refinement network.}

After the two-stage training of geometry and appearance, our \method model can directly generate multi-view GS-maps, efficiently representing the 3DGS objects. 
To further enhance the reconstruction quality, we introduce a U-Net-based refinement network to refine the GS-maps.
We follow the way in MASt3R \cite{mast3r_eccv24} to make the local stereo model to predict matching features intrinsically contain rich geometric information from the pair-wise images.
Such geometric information can serve as a prior for multi-view correspondence learning during the refinement network training.
In addition, camera pose information is often available in synthetic data or multi-view generative models such as Mvdream \cite{shi2023mvdream} or V3D\cite{chen2024v3d}, which can also be leveraged by our refinement network to produce more fine-grained results.

In specific, we directly concatenate the multi-view GS-maps with the matching features and RGB images as the input to the refinement network.
Additionally, camera poses can be encoded as Plücker ray embeddings \cite{xu2023dmv3d} and optionally concatenated with the multi-view inputs.
The refinement network is a U-Net network and constructed with residual layers \cite{he2016deep} and self-attention layers \cite{vaswani2017attention} following previous works \cite{ho2020denoising, metzer2023latent, szymanowicz2024splatter}. 
Self-attention layers are incorporated at the deeper layers where the input multi-view feature embedding resolution is down-sampled.
We flatten and concatenate the multi-view features before applying self-attention layers, facilitating information propagation across multiple views, similar to prior multi-view diffusion models \cite{wang2023imagedream, shi2023mvdream}. The output of the refinement network is added to the input Gaussian maps as a residual connection.


\subsection{Training}

Our training contains two stages to optimize the predicted 3DGS geometry and appearance. 
The training is conducted on the large-scale 3D object \cite{deitke2023objaverse} dataset for generalizability and robustness.
In the first stage, we focus on training the \method model with the point prediction head for geometry prediction.
To ensure accurate ground truth geometry for training, we randomly sample 3D points from the surface of the 3D object to construct a ground truth point cloud $\hat{S}$. This approach provides a reliable reference for learning precise multi-view point-maps.
During training, we employ a regression-based method to optimize the prediction of multi-view point-maps. 
Specifically, we randomly sample points from the predicted point-maps to form a predicted point cloud $S$. To measure the alignment between the predicted and ground truth point clouds, we use the Chamfer Distance loss:
\begin{equation}
    \mathcal{L}_{\text{Chamfer}}=\frac{1}{\left|S\right|} \sum_{x \in S} \min _{y \in \hat{S}}\|x-y\|_2^2+\frac{1}{\left|\hat{S}\right|} \sum_{y \in \hat{S}} \min _{x \in S}\|x-y\|_2^2,
\end{equation}
where $S$ is constructed by randomly sampling $10k$ points from the multi-view point maps $P$ within the valid foreground area mask $M$.

We further optimize the geometry by incorporating depth values $D$. The predicted depth $D$ is derived by transforming the predicted point maps $P$ from the world coordinate system to the camera coordinate system, using ground truth camera parameters. These camera parameters are utilized only during training.

To ensure accurate and smooth depth predictions, we apply two loss functions: an $\ell_1$ loss to minimize the absolute difference between the predicted depth $D$ and the ground truth depth $\hat{D}$, and a gradient loss to preserve structural details and edge consistency: 
\begin{equation}
  \begin{aligned}
  & \mathcal{L}_{\text{depth}}=\alpha \cdot|D-\hat{D}| + \\
  & \beta \cdot(|\partial_x D-\partial_x \hat{D}|+|\partial_y D-\partial_y \hat{D}|), \\
   \end{aligned}
\end{equation}
where $\partial_x$ and $\partial_y$ denotes the gradients on the $x$ and $y$ directions.
Following UniMatch \cite{xu2023unifying}, we use $\alpha=20$ and $\beta=20$.

In the second stage, we train our \method model with the Gaussian prediction head and refinement network for appearance prediction and novel view synthesis.
We train our full model with a combination of mean squared error (MSE) and LPIPS \cite{Zhang_Isola_Efros_Shechtman_Wang_2018} losses between rendered images and ground truth:
\begin{equation}
\mathcal{L}_{\mathrm{rgb}}=\operatorname{MSE}(I_{rgb}, \hat{I}_{rgb})+\lambda \cdot \operatorname{LPIPS}(I_{rgb}, \hat{I}_{rgb}),
\end{equation}
where $I_{rgb}$ is the rendered image, $\hat{I}_{rgb}$ is the ground truth, and LPIPS loss weight $\lambda$ is set to 0.05.
We also apply MSE loss on the predicted alpha mask:
\begin{equation}
\mathcal{L}_\alpha=\operatorname{MSE}(I_\alpha, \hat{I}_\alpha),
\end{equation}
where $I_\alpha$ is the predicted alpha image and $\hat{I}_\alpha$ is the ground truth alpha.



\subsection{Implementation}
We train our model on 8 NVIDIA V100 (32G) GPUs for about 2 days. 
Previous methods \cite{tang2024lgm, hong2023lrm, xu2024grm} which requires higher-end A100 (80G) GPUs for more than \textbf{3000 hours}.
Our method only needs lower-end V100 GPUs for slightly above \textbf{300 hours}, \textit{which significantly reduces the training hours to \textbf{1/10} even with lower-end GPUs.}
Each GPU uses a batch size of 6 with bfloat16 precision, resulting in a total batch size of 48. 
We set 4 images as the input images by default and randomly select 8 camera views in each batch.
The AdamW \cite{loshchilov2017decoupled} optimizer is adopted with the learning rate of $4 \times 10^{-4}$, weight decay of $0.05$, and betas of $(0.9,0.95)$.

\begin{figure*}[t]
  \centering
  \begin{subfigure}{\linewidth}
    \includegraphics[width=1.0\linewidth]{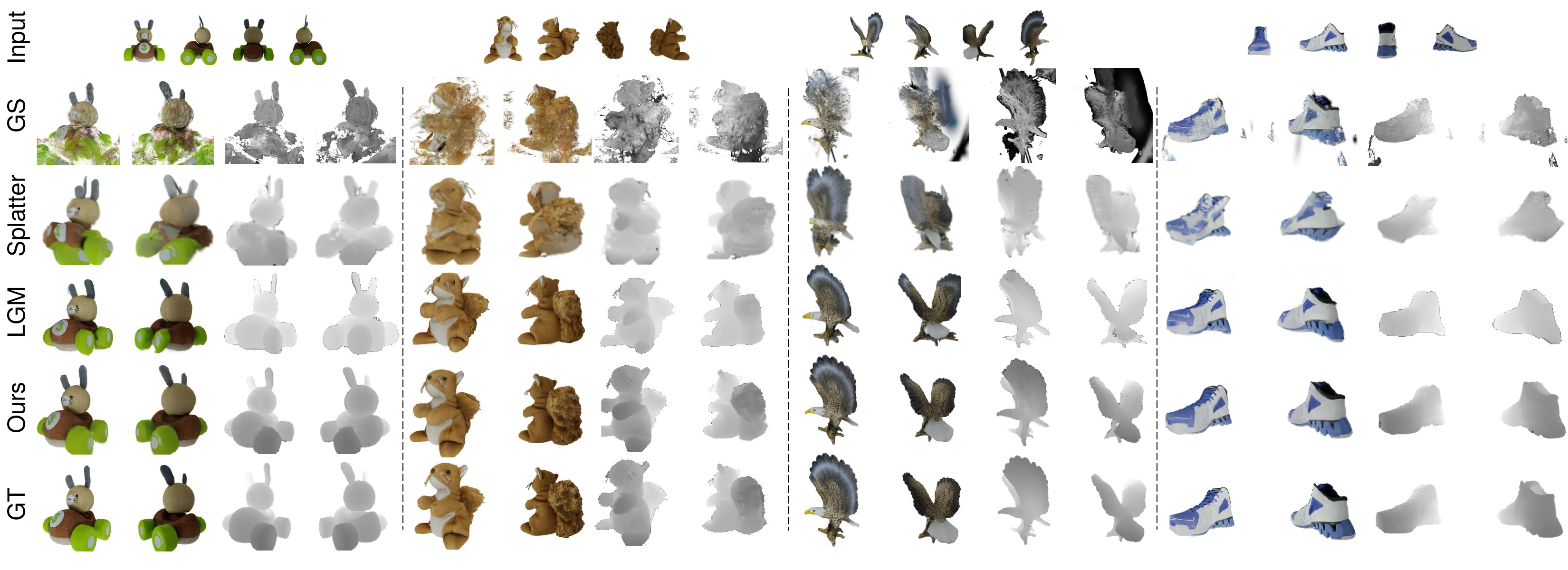}
  \end{subfigure}
  
  \caption{\textbf{Multi-view reconstruction.} Given the same multi-view inputs, the standard GS \cite{kerbl3Dgaussians} totally fails to render novel view images. 
  SplatterImage hardly reconstructs the 3D objects under the multiple view images. 
  Although LGM \cite{tang2024lgm} can reconstruct finer geometric structures and appearance details, it still faces challenges in maintaining consistency and avoiding artifacts.
  Our method can generate both high quality geometry and appearance for the 3D objects.}
  \label{fig:sparse_view_results}
\end{figure*}

\section{Experiments}
\label{sec:experiments}

In this section we introduce our experiments of the details about datasets, training and evaluation.
As our model is essentially a multi-view reconstruction model, we evaluate the 3D reconstruction performance with multi-view images.
We also combine the multi-view diffusion model with our model to evaluate the performance for single image-to-3D generation.

\paragraph{Dataset.}

Our training dataset consists of multi-view images rendered from the Objaverse \cite{deitke2023objaverse} dataset. Unlike existing methods that typically require a large-scale dataset of the full set of Objaverse with around 80K objects \cite{hong2023lrm}, or require around one million rendering views \cite{tang2024lgm}.
We utilize the LVIS annotation to obtain a subset of Objaverse \cite{deitke2023objaverse}, which contains around $45$k high-quality objects.
For each object in the dataset, we render 512 × 512 images and depth maps from 32 random viewpoints. 
Our datasets contain around 144k rendering views in total, which is sufficient for our model training to obtain superior performance and reduce the reliance on large-scale datasets.
We evaluate the qualitative and quantitative performance using Google Scanned Objects (GSO) \cite{downs2022google} datasets.
The GSO dataset comprises around 1,000 objects, from which we randomly select 100 objects for our evaluation set. 
Specifically, we render 16 images of each object in the GSO evaluation set in an orbiting trajectory with uniform azimuths varying positive elevations in $\{0^{\circ}, 20^{\circ}\}$ for sampling on the top semi-sphere of an object.

\subsection{Results on multi-view reconstruction} 

We compare our method with Gaussian Splatting \cite{kerbl3Dgaussians}, SplatterImage \cite{szymanowicz2024splatter} and LGM \cite{tang2024lgm} on the GSO dataset.
For all baselines, we use four input views with camera elevation angle = 0 and azimuths degrees = $\{0,90,180,270\}$ to cover the entire object and evaluate the reconstruction quality on the remaining 12 views.
SplatterImage \cite{szymanowicz2024splatter} was originally designed to handle multi-view reconstruction using only two input views. When all four input view images are used, it results in corrupted reconstructions. Therefore, to align with the original implementation, we only use the first two views from the input images.

We use PSNR, SSIM, and LPIPS \cite{Zhang_Isola_Efros_Shechtman_Wang_2018} to measure the appearance reconstruction quality.
Following previous methods \cite{tang2018ba, xu2023unifying}, we use Absolute Relative difference (Abs Rel), Squared Relative difference (Sq Rel), and Root Mean Squared Error (RMSE) for evaluation of the rendered depth maps to measure geometry reconstruction quality.
All comparisons are conducted at a resolution of $256$.
As shown in \Fref{fig:sparse_view_results}, our reconstructions accurately represent the geometric structures without any visible floaters and capture appearance details with higher quality than the baseline methods.
\Tref{tab:sparse_view_appearance} and \Tref{tab:sparse_view_geometry} demonstrate the quantitative reconstruction results, our method outperforms all baselines across both appearance and geometry metrics with obvious margins. 
Though our method takes a slightly higher inference speed than LGM \cite{tang2024lgm}, it shows a balance between high reconstruction quality and acceptable inference efficiency.
We provide more results for real-world objects and scene scenarios in the supplementary material to further demonstrate the scalability of our method.

\begin{table}[h]
\centering
\caption{Appearance reconstruction quality evaluation for multi-view reconstruction on Google Scanned Objects datasets~\cite{downs2022google}.}
\label{tab:sparse_view_appearance}

\begin{adjustbox}{width=0.475\textwidth}
\begin{tabular}{l|cccccc}
\toprule
Methods & PSNR\textuparrow & SSIM\textuparrow & LPIPS\textdownarrow & Time \\
\midrule
GS~\cite{kerbl3Dgaussians}                    & 17.81 & 0.6946  & 0.2948 & 270 sec \\
SplatterImage~\cite{szymanowicz2024splatter}  & 18.05 & 0.8221  & 0.1764 & \textbf{0.32 sec} \\
LGM~\cite{tang2024lgm}                        & 23.79 & 0.8791  & 0.1035 & 1.67 sec \\
Ours &  \textbf{27.12}                        & \textbf{0.9352} & \textbf{0.0534} & 2.62 sec \\
\bottomrule
\end{tabular}
\end{adjustbox}
\end{table}
\begin{table}[h]
\centering
\caption{Geometry reconstruction quality evaluation for multi-view reconstruction on Google Scanned Objects datasets~\cite{downs2022google}.}
\label{tab:sparse_view_geometry}

\begin{adjustbox}{width=0.425\textwidth}
\begin{tabular}{l|cccccc}
\toprule
Method & Abs Rel\textdownarrow & Sq Rel\textdownarrow & RMSE\textdownarrow \\
\midrule
GS~\cite{kerbl3Dgaussians}                       & 0.4570 & 0.1982 & 0.5031 \\
SplatterImage~\cite{szymanowicz2024splatter}     & 0.3223 & 0.1178 & 0.3779 \\
LGM~\cite{tang2024lgm}                           & 0.2695 & 0.0758 & 0.1973 \\
Ours                                             & \textbf{0.1112} & \textbf{0.0238} & \textbf{0.1175} \\
\bottomrule
\end{tabular}
\end{adjustbox}
\end{table}

\begin{figure*}[tp]
  \centering
  \begin{subfigure}{\linewidth}
    \includegraphics[width=1.0\linewidth]{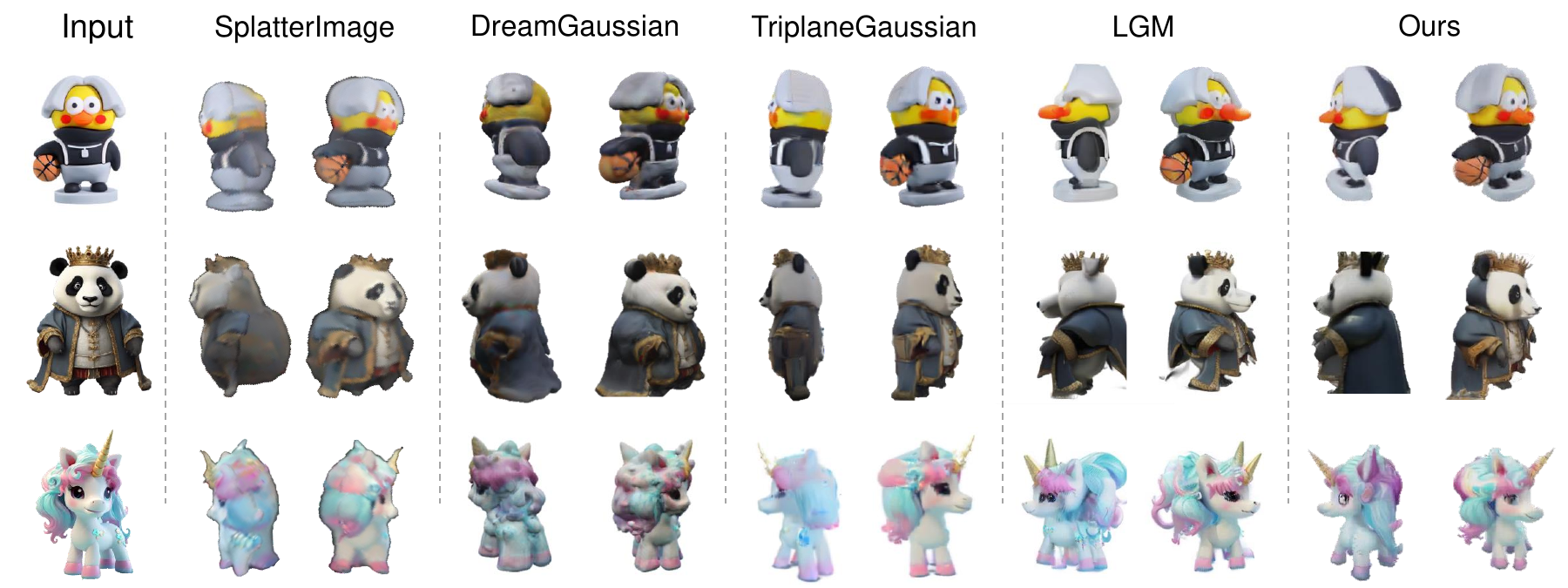}
  \end{subfigure}
  
  \caption{\textbf{Single Image-to-3D generation}. Our method generates the 3D object with better visual quality and more consistent geometry than the baseline methods.}
  \label{fig:single-image-to-3d}
\end{figure*}

\begin{figure*}[tp]
  \centering
  \begin{subfigure}{\linewidth}
    \includegraphics[width=1.0\linewidth]{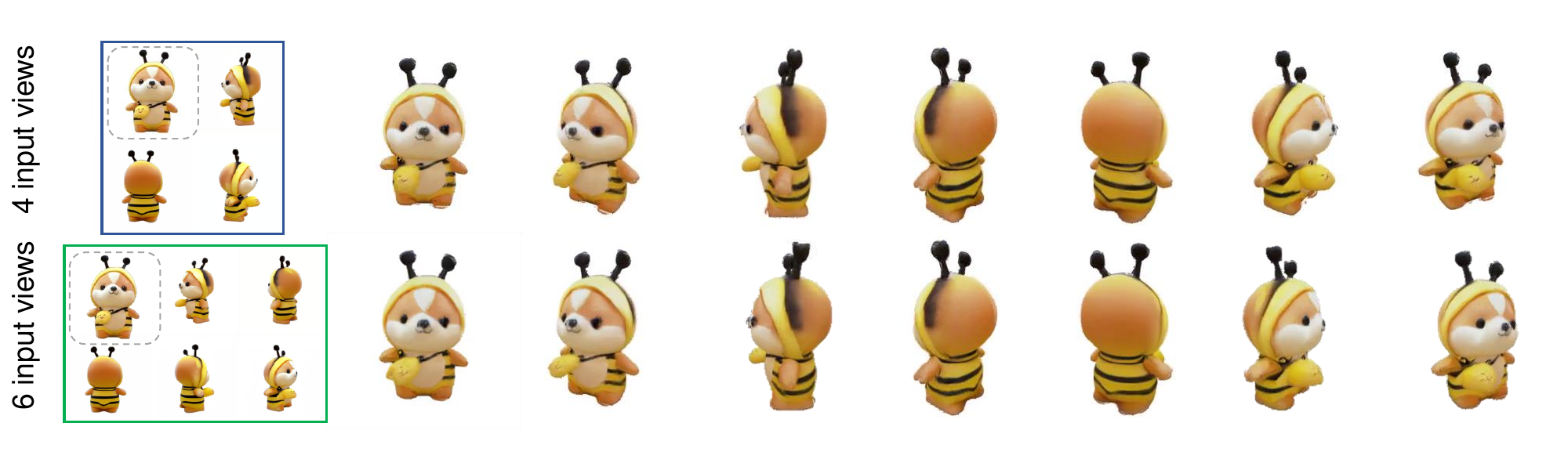}
  \end{subfigure}
  
  \caption{Single Image-to-3D generation with 4 input views and 6 input views. We use V3D \cite{chen2024v3d} to generate multi-view images, with the input image scaled up for better visualization and positioned in the upper-left corner of each input views group.}
  \label{fig:number_views}
\end{figure*}

\subsection{Results on single Image-to-3D generation} 



We utilize the recently proposed V3D \cite{chen2024v3d} as our multi-view diffusion model since its superior performance in generating high-quality multi-view consistent images as the multi-view input for our model.
The camera elevation is fixed to 0, and azimuths are set to $[0,80,180,280]$ degree for the four input views condition and $[0,60,120,180,240,300]$ degree for the six input views condition.
The single input image is preprocessed for background removal using the $\operatorname{U^2Net}$ \cite{Qin_2020_PR} method.

We compare our method with four recent baselines \cite{szymanowicz2024splatter, tang2023dreamgaussian, zou2024triplane, tang2024lgm} that are capable of generating 3D Gaussians from a single image input.
As shown in \Fref{fig:single-image-to-3d}, we display the image rendered by different methods for comparison. 
SplatterImage \cite{szymanowicz2024splatter} shows poor novel view synthesis results. 
DreamGaussian \cite{tang2023dreamgaussian} is an optimization-based method and shows various geometry artifacts.
TriplaneGaussian \cite{zou2024triplane} learns to generate flat geometry, leading to unrealistic results and missing content details.
LGM \cite{tang2024lgm} shows multi-view inconsistency, resulting in geometry artifacts and blurry texture.
Our method generates consistent results with higher reconstruction quality in both appearance and geometry.
We also quantitatively evaluate the reconstruction result on the GSO data by using PSNR, SSIM, and LPIPS \cite{Zhang_Isola_Efros_Shechtman_Wang_2018} metrics as shown in \Tref{tab:image-to-3d}.

\begin{table}[thb!]
\centering
\caption{\textbf{Single image-to-3D generation on GSO dataset.} Combined with an image-to-multiview diffusion model~\cite{shi2023zero123++}, our method can be used for single image-to-3D generation.}
\label{tab:image-to-3d}
\begin{adjustbox}{width=0.425\textwidth}
\begin{tabular}{l|ccccccc}
\toprule
Method              & PSNR\textuparrow & SSIM\textuparrow & LPIPS\textdownarrow \\ 
\midrule
DreamGaussian~\cite{tang2023dreamgaussian}    & 18.23 & 0.8161 & 0.1897 \\
\hline
Splatter-Image~\cite{szymanowicz2024splatter} & 16.06 & 0.7804 & 0.3289 \\
TriplaneGaussian~\cite{zou2024triplane}       & 16.48 & 0.8085 & 0.2567 \\ 
LGM~\cite{tang2024lgm}                        & 17.04 & 0.8173 & 0.2274 \\
Ours               & \textbf{19.31} & \textbf{0.8238} & \textbf{0.1412} \\
\bottomrule
\end{tabular}
\end{adjustbox}
\end{table}




\subsection{Ablation study}

\paragraph{Number of Views.} 
Besides the 4 input image settings, we also train our model with 6 input images and generate 3D Gaussians under the supervision of 8 images.
As illustrated in \Fref{fig:number_views}, our method achieves faithful 3DGS reconstruction using both 4 and 6 input images. Notably, qualitative results with 6 input images demonstrate greater consistency, indicating that leveraging additional images enhances reconstruction quality.
Previous methods typically use a fixed number of input images \cite{szymanowicz2024splatter, tang2024lgm}. 
In contrast, our model can reconstruct 3D objects with a larger number of input images, allowing more synthetic images as input to enhance the overall reconstruction quality.
We provide more details about the qualitative and quantitative results in the supplementary materials.

\paragraph{Training time efficiency.}

\begin{figure}[tp]
  \centering
  \begin{subfigure}{\linewidth}
    \includegraphics[width=1.0\linewidth]{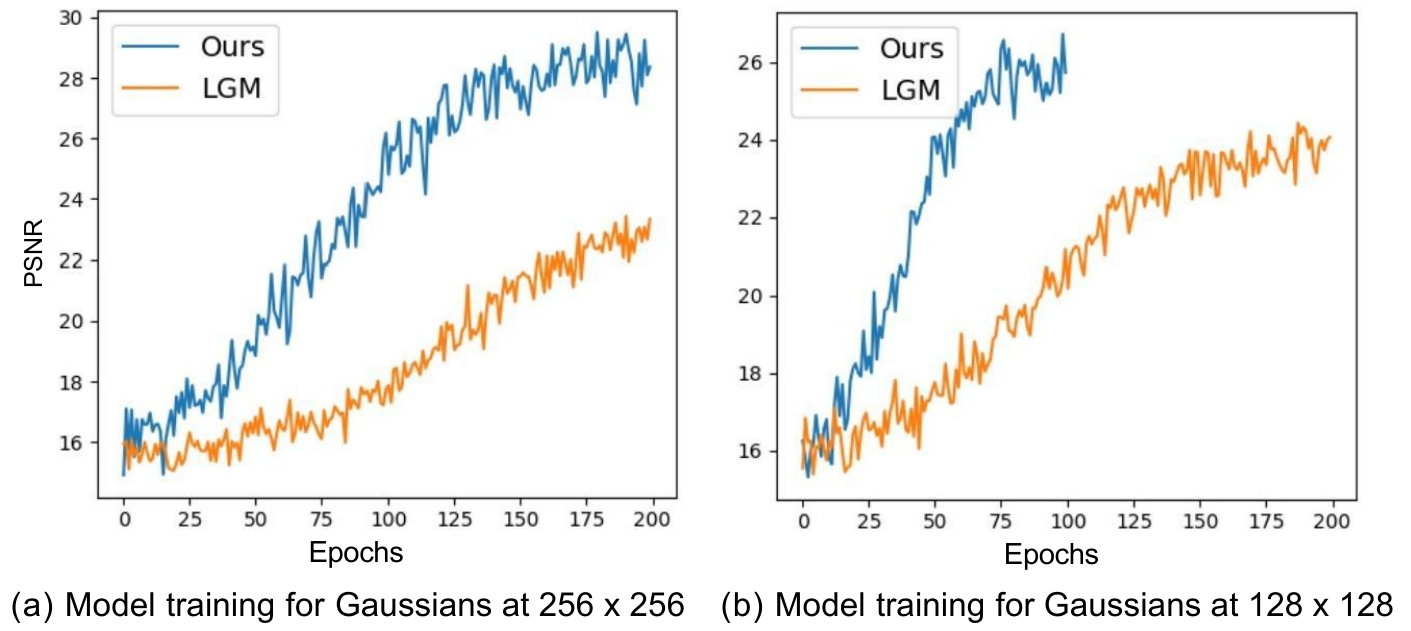}
  \end{subfigure}
  
  \caption{\textbf{Training time efficiency.} Comparison of model training time for predicting Gaussian features at the resolution of (a) $256 \times 256$ and (b) $128 \times 128$ as the output 3D Gaussians for rendering.}
  \label{fig:training_time}
\end{figure}

We compare the training efficiency with LGM \cite{tang2024lgm}. 
Our experiments are conducted using our default training resources, which include 8 NVIDIA V100 (32GB) GPUs, and we train on the Objaverse \cite{deitke2023objaverse} LVIS subset.
To make a fair comparison, we set the input to 4 images and set the output Gaussians resolution to $256 \times 256 = 65536$ for both methods.
Specifically, we add one more upsampling layer for LGM to match the input resolution.
As shown in \Fref{fig:training_time} (a), our method converges within 200 epochs. In contrast, LGM \cite{tang2024lgm} still requires significantly more training time to reach convergence.

Additionally, we experimented with a smaller output Gaussian resolution of 
$128 \times 128 = 16384$, which follows the original asymmetric design of LGM \cite{tang2024lgm}. 
The smaller size of predicted 3D Gaussians enables a more effective batch size, allowing our method to converge faster within around 100 epochs.
Meanwhile, LGM\cite{tang2024lgm} still requires more than 200 epochs to complete training.
Both experiments indicate that our method successfully utilizes the disentangled framework to accelerate the reconstruction model regression.

\paragraph{Model components.}

\begin{figure}[t]
  \centering
  \begin{subfigure}{\linewidth}
    \includegraphics[width=1.0\linewidth]{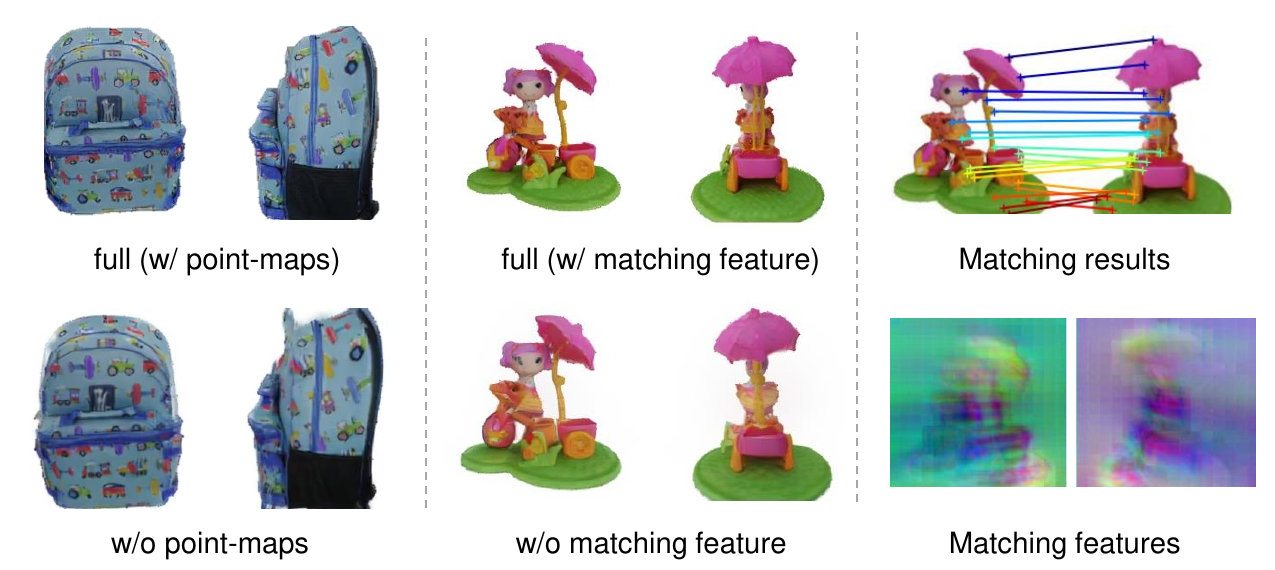}
  \end{subfigure}
  
  \caption{\textbf{Qualitative ablation study for model components.} Without point prediction head, the 3D object can exhibit incorrect geometry and lower reconstruction quality. Without matching features, the refinement network only generate blur results, while including matching features can accelerate the regression process and display clearer results.}
  \label{fig:ablation_model_component}
\end{figure}

We ablate different components of our full model in \Tref{tab:ablation_model_component}.
We ablate the point prediction head and make the Gaussian prediction head to predict the full 3DGS parameters.
We also remove the refinement network and ablate its matching prior as input.
To make the ablation study more efficient, we conduct all experiments with 4 input images, which can be finished in 100 epochs.
We further visualize the results as shown in \Fref{fig:ablation_model_component}.
Removing point prediction head can lead to incorrect geometry reconstruction, potentially rendering some areas as flat regions.
Excluding the matching feature prior results in a blurrier appearance reconstruction compared to the full model.
We visualize the matching prior containing correspondence information across views, demonstrating its benefits for improving performance and efficiency for refinement network.

\begin{table}[h]

\centering
\caption{\textbf{Ablation study for model components.}}
\label{tab:ablation_model_component}

\begin{adjustbox}{width=0.475\textwidth}
\begin{tabular}{l|cccccc}
\toprule
Setting & PSNR\textuparrow & SSIM\textuparrow & Abs Rel\textdownarrow & RMSE\textdownarrow \\
\midrule
full                     & \textbf{25.32} & \textbf{0.9347} & \textbf{0.1029} & \textbf{0.1039} \\
w/o refinement network   & 24.72 & 0.9230 & 0.1053 & 0.1067 \\
w/o point-maps           & 22.15 & 0.9044 & 0.1191 & 0.1199 \\
w/o matching feature     & 20.39 & 0.8555 & 0.1314 & 0.1261 \\
\bottomrule
\end{tabular}
\end{adjustbox}
\end{table}



\section{Conclusion}

In this work, we introduce \method, a novel approach that leverages stereo features from local image pairs, and fuses them through global attention blocks to predict multi-view point-maps and Gaussian features. 
Our method effectively disentangles 3D Gaussian prediction into geometry and appearance components to achieve pose-free and generalizable 3D Gaussian reconstruction. It provides greater flexibility for real-world applications and accelerates the training process for high-quality 3DGS content generation.


\section*{Acknowledgement}
Renjie Group is supported by the National Natural Science Foundation of China under Grant No. 62302415, Guangdong Basic and Applied Basic Research Foundation under Grant No. 2022A1515110692, 2024A1515012822. 
This work was also supported in part by the National Natural Science Foundation of China Grant 62471278, in part by the Taishan Scholar Project of Shandong Province under Grant tsqn202306079.

\bibliographystyle{ACM-Reference-Format}
\bibliography{main}

\clearpage
\setcounter{page}{1}
\section*{Supplementary Material}



In this supplementary material, we first discuss multi-view reconstruction tasks for real-world scenes to demonstrate the scalability of our method.
We then explain more details of our model's training and inference efficiency.
We also dive deep to discuss the effects of increasing the number of views to enhance our model's performance. 
We provide a comprehensive evaluation of both Single-Image-to-3D and multi-view reconstruction scenarios.


\section{Real world scene}



\begin{figure}[htbp]
   \vspace{-10pt}
   \centering
   \includegraphics[width=1.0\linewidth]{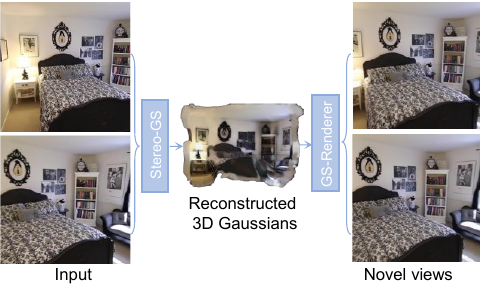}
   \caption{Extend Stereo-GS on real-world scenarios.}
   \vspace{-10pt}
   \label{fig:re10k_qualitative}
\end{figure}

Our framework is also applicable to real-world scenes. 
As shown in the \Fref{fig:re10k_qualitative}, Stereo-GS can also be extended into real-world scenarios by training and evaluation on the RealEstate10K dataset (Re10K). 
Specifically, due to the unbounded nature of real-world scenes, we anchor the predicted points-maps in the canonical coordinate system of the first view. 
In this setup, the first view serves as the origin of the camera coordinate system, and all other views are referenced relative to the position of the first view's camera.

Our disentangled geometry and appearance training framework is still applicable to real-world scene scenarios. 
In particular, for the Re10K dataset, while ground truth sparse point clouds are provided, ground truth depth maps are unavailable. 
Consequently, in the first stage of training for geometry, we optimize point predictions using only the Chamfer distance loss, which measures the discrepancy between the predicted point cloud and the ground truth point cloud.

In the second stage, we train the Gaussian prediction head to predict Gaussian features. 
These features are then combined with point maps to generate GS-maps, under the supervision of the ground truth images.
We continue to apply the photometric loss in RGB space and the perceptual loss for appearance optimization.
As shown in the \Tref{table:re10k_quantitative}, comparing with previous PixelSplat \cite{charatan2024pixelsplat} and MVSplat \cite{chen2024mvsplat} baselines, 
our approach outperforms previous baselines, including PixelSplat \cite{charatan2024pixelsplat} and MVSplat \cite{chen2024mvsplat}, achieving superior results on the RE10K dataset across various input-view settings. These results further validate the scalability of our method.

\begin{table}[htbp]
\centering
\caption{
Quantitative evaluation on RealEstate10K dataset.
}
\label{table:re10k_quantitative}
\begin{adjustbox}{width=0.45\textwidth}
\begin{tabular}{c|cc|cc|cc}
\toprule
Method & \multicolumn{2}{c|}{PixelSplat} & \multicolumn{2}{c|}{MVSplat} & \multicolumn{2}{c}{Ours} \\
Metric & PSNR & SSIM & PSNR & SSIM  & PSNR & SSIM  \\
\midrule
2 views & 25.89  & 0.858 & 26.39 & 0.869 & 26.67 & 0.875 \\ 
4 views & N.A.  & N.A. & 26.54 & 0.867 & 27.12 & 0.883 \\ 

\bottomrule
\end{tabular}
\end{adjustbox}
\vspace{-5pt}
\end{table}

\vspace{-10pt}
\section{Training and inference efficiency}
\label{sec:rationale}

We conduct a comprehensive evaluation of both the training and inference efficiency of our method.
We estimate the inference speeds at the image resolution of 512 for the multi-view reconstruction tasks, as shown in \Tref{tab:sparse_view_time}.
Thanks to the high efficiency of the stereo feature backbone which only takes 1.73 seconds for 4 input views and 3.04 seconds for 6 input views, our model can achieve fast 3D reconstruction. It only takes 2.62 seconds for 4 input views and 4.16 seconds for 6 input views.
This speed remains efficient for multi-view reconstructions while achieving higher-quality reconstruction.
For the Single Image-to-3D tasks, our diffusion model baseline \cite{chen2024v3d}, takes around 45 seconds to generate 18 frames of multi-view images. 
Thus, our pipeline can complete the Single Image-to-3D reconstruction within one minute, delivering better results than previous baselines \cite{szymanowicz2024splatter, zou2024triplane, tang2024lgm} while maintaining high efficiency.

\begin{table}[h]
\centering
\caption{Inference time of different components in our model with 4 and 6 input-view images at the resolution of 512.}
\label{tab:sparse_view_time}

\begin{adjustbox}{width=0.425\textwidth}
\begin{tabular}{l|cccccc}
\toprule
Methods & Stereo-GS & Refinement network & Full \\
\midrule
Ours w/ 4 views & 1.73 sec & 0.89 sec & 2.62 sec \\
Ours w/ 6 views & 3.04 sec & 1.12 sec & 4.16 sec \\
\bottomrule
\end{tabular}
\end{adjustbox}
\end{table}


Regarding training efficiency, utilizing 8 V100 (32GB) GPUs can still be resource-intensive in some limited training scenarios. To address this, we conduct an experiment using \textbf{1 NVIDIA V100 (32GB) GPUs}, training for 200 epochs on a smaller model with 4 input views that at the resolution of $256 \times 256$ for the output GS-maps, and then fine-tuning on 6 input views for 50 epochs. 
This training process can be completed within \textbf{approximately 7 days}, demonstrating that our method is still efficient and scalable for most use cases. This makes it feasible for many users to train their own 3D Gaussian reconstruction models without requiring excessive resources.

\section{Number of views}
\label{sec:supp_num_views}

To train our model for 6 input views, we fine-tune our model on the previous model which is training with 4 input view images.
During training, we still randomly select 8 camera views for each object and set the first 6 images as the input images.
Since the larger numbers of input view images, each GPU can use a batch size of 4 with bfloat16 precision, resulting in a total batch size of 32. 
We use the same learning rate setting with the training for 4 views and the training can be finished in 50 epochs. 

\paragraph{Single Image-to-3D.}

\begin{figure*}[htbp]
  \centering
  \begin{subfigure}{\linewidth}
    \includegraphics[width=1.0\linewidth]{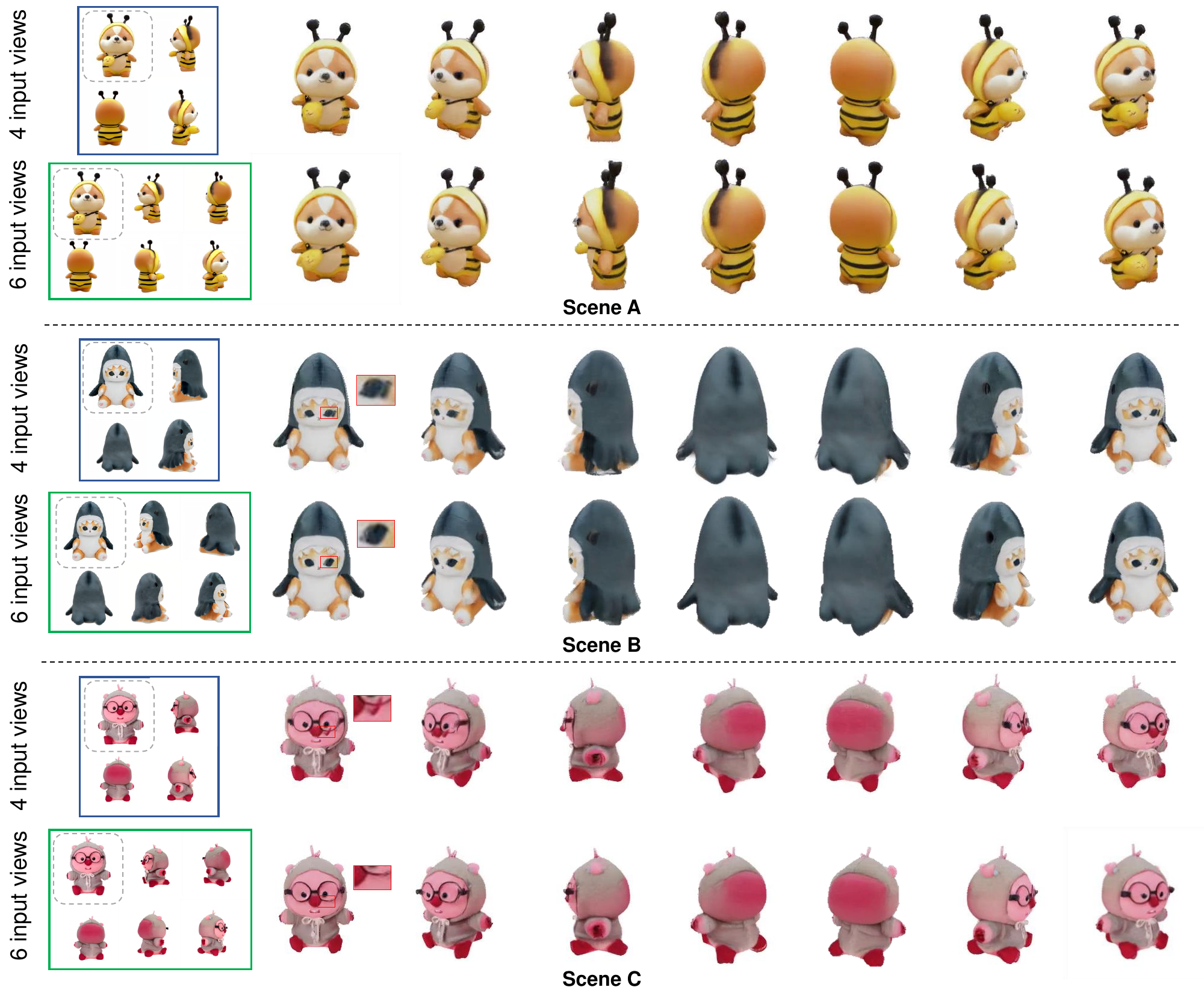}
  \end{subfigure}
  
  \caption{\textbf{Single Image-to-3D generation with 4 input views and 6 input views.} We use V3D \cite{chen2024v3d} to generate multi-view images, with the input image scaled up for better visualization and positioned in the upper-left corner of each input views group.}
  \label{fig:single-image-to-3d-supp}
\end{figure*}

Based on the multi-view synthesis capability of the video diffusion model V3D \cite{chen2024v3d}, we can generate multi-view images from a single image input. 
Different from the multi-view diffusion models \cite{shi2023mvdream, wang2023imagedream} can only generate a few output views (typically 4 images), V3D \cite{chen2024v3d} can generate 18 images. It enhances the reconstruction quality by providing more geometry and appearance information.
Our default model utilizes 4 images from the V3D \cite{chen2024v3d} synthetic results at an elevation of 0 degrees and azimuths of $[0, 80, 180, 280]$ degrees. Additionally, our fine-tuned model samples 6 images from the synthetic samples at an elevation of 0 degrees and azimuths of $[0, 60, 120, 180, 240, 300]$ degrees.
We display the qualitative examples in the \Fref{fig:single-image-to-3d-supp}. As shown in \Fref{fig:single-image-to-3d-supp} scene A, both 4 input views and 6 input views can generate promising reconstruction results. 
Our fine-tuned model with 6 input views can generate clearer results due to the increased number of 3D Gaussians.

However, the reconstruction model may face challenges in certain scenarios.
As shown in \Fref{fig:single-image-to-3d-supp} scene B and C, the reconstruction results from our default 4-input-views model may exhibit artifacts.
These artifacts are primarily due to slight inconsistencies in the synthetic images.
When reconstructing 3D objects from images with large azimuth gaps (90 degrees between 4 input images), misalignments can occur, particularly in regions such as the eyes or nose.
To address this, increasing the number of input view images reduces the azimuth gaps, leading to greater overlap and enhanced consistency among the multi-view images.
By leveraging 6 synthetic images, our fine-tuned model effectively mitigates misalignment artifacts, resulting in clearer and more accurate 3D reconstructions compared to the 4-input-views model.

\paragraph{Multi-view reconstruction.}

We discuss both qualitative and quantitative results when increasing the input number of views from 4 images to 6 images. 
For the 4-input-views condition, we maintain the same camera pose setting with elevations of 0 and azimuths at $[0, 90, 180, 270]$ degrees. 
For the 6-input-views condition, we set the camera pose with elevations of 0 and azimuths of $[0, 60, 120, 180, 240, 300]$ degrees. 
The reconstruction quality is still evaluated at azimuths of $[45, 135, 225, 315]$.

\begin{figure*}[htbp]
  \centering
  \begin{subfigure}{\linewidth}
    \includegraphics[width=1.0\linewidth]{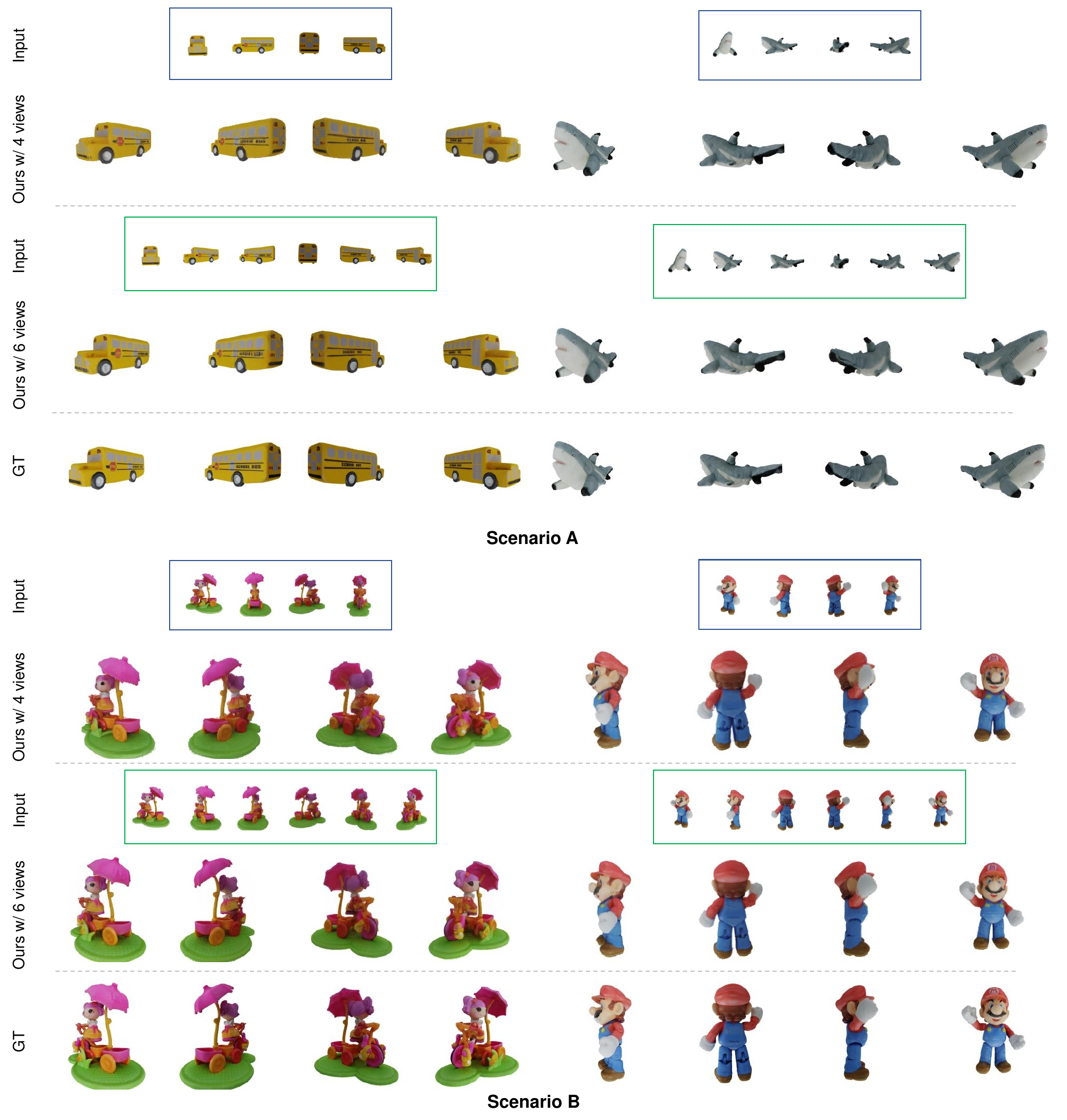}
  \end{subfigure}
  
  \caption{\textbf{Multi-view reconstruction with 4 input views and 6 input views.} In Scenario A which contains 3D objects with simpler geometry, our model can accurately reconstruct the 3D objects using both 4 input views and 6 input views. While Scenario B contains 3D objects with more complex geometry, our model with 6 input views can achieve superior results compared with the 4 input views condition.}
  \label{fig:sparse_view_results_supp}
\end{figure*}

We demonstrate qualitative results in \Fref{fig:sparse_view_results_supp}. 
With accurate camera poses and consistent input images, both our models for 4-input-views and 6-input-views can faithfully reconstruct the 3D objects in Scenario A, which contains simpler geometry.
Meanwhile, we can observe an improvement in the appearance reconstruction quality for the 6‐input‐views model in \Fref{fig:sparse_view_results_supp} Scenario B, which contains more complex geometry.
By using 6 input views with more information and reduced azimuth gaps, our model can generate more Gaussians and achieve a precise 3D reconstruction.
Quantitative results presented in \Tref{tab:sparse_view_appearance_supp} indicate an improvement across all three metrics (PSNR, SSIM, LPIPS) for the 6‐input‐views condition.

\begin{table}[h]
\centering
\caption{Appearance reconstruction quality evaluation for sparse-view reconstruction on Google Scanned Objects datasets~\cite{downs2022google}.}
\label{tab:sparse_view_appearance_supp}

\begin{adjustbox}{width=0.41\textwidth}
\begin{tabular}{l|cccccc}
\toprule
Methods & PSNR\textuparrow & SSIM\textuparrow & LPIPS\textdownarrow \\
\midrule
Ours w/ 4 views &  27.12 & 0.9352 & 0.0534 \\
Ours w/ 6 views &  28.20 & 0.9439 & 0.0522 \\
\bottomrule
\end{tabular}
\end{adjustbox}
\end{table}

\end{document}